\documentclass[10pt]{article}

\pdfoutput=1 
\usepackage{lipsum} 

\usepackage{float} 
\usepackage[sc]{mathpazo} 
\usepackage[T1]{fontenc} 
\usepackage{microtype} 

\usepackage{multicol} 
\usepackage[hang, small,labelfont=bf,up,textfont=it,up]{caption} 
\usepackage{booktabs} 
\usepackage{float} 
\usepackage[pdftex,breaklinks,colorlinks,
citecolor=blue,
urlcolor=blue]{hyperref}

\usepackage{fancyvrb}
\fvset{fontsize=\normalsize}

\usepackage{lettrine} 
\usepackage{paralist} 

\usepackage{abstract} 

\usepackage{titlesec} 
\renewcommand\thesection{\Roman{section}}
\titleformat{\section}[block]{\large\scshape\centering}{\thesection.}{1em}{} 

\usepackage[top=0.75in]{geometry}

\newcommand{\OMIT}[1]{}

\usepackage{color}
\usepackage[pdftex]{graphicx}
\usepackage{amsmath}
\usepackage{algpseudocode}
\usepackage{bm,nicefrac}
\usepackage{upgreek}
\usepackage[bb=boondox]{mathalfa}

\begin{document}
\thispagestyle{empty} 

%
\vspace*{9ex}
\begin{center}
{\large\bf 
On Uncensored Mean First-Passage-Time Performance Experiments \\[0.5ex]
with Multiwalk in $\mathbb{R}^p$: a New Stochastic Optimization Algorithm
}
\par\vspace*{2ex}

\begin{figure}[h!] 
\hspace*{4em}  
\begin{minipage}{0.30\textwidth}
\centering
Franc Brglez\\   
Computer Science, \\
NC State University\\ 
Raleigh, NC 27695, USA\\
\href{mailto:brglez@ncsu.edu}{brglez@ncsu.edu}  
\end{minipage}
\hspace*{-2em}
\begin{minipage}{0.60\textwidth}
\centering 
\includegraphics[width=0.60\textwidth]{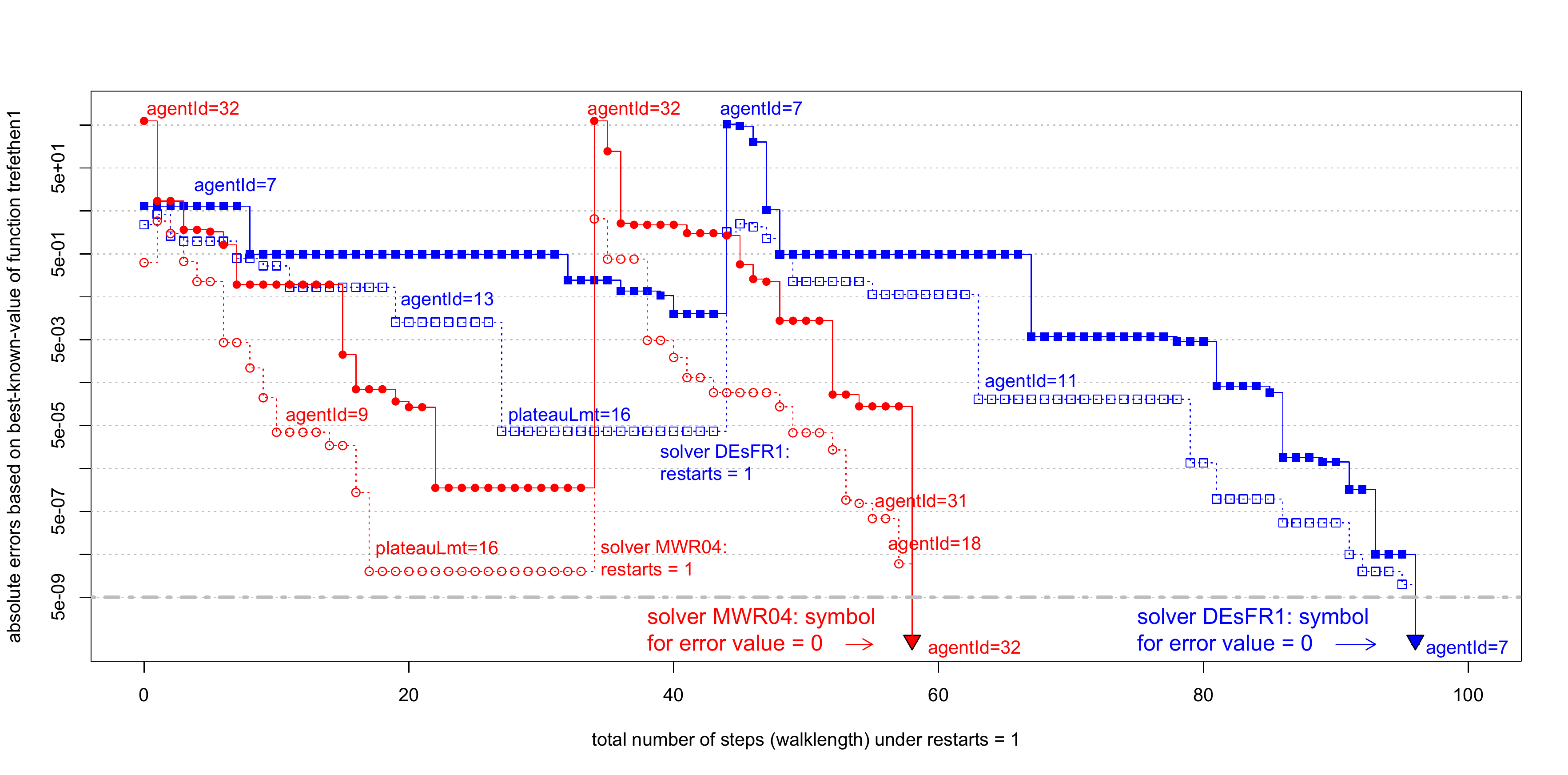} 
\end{minipage}
\end{figure}
\end{center}

\begin{center}
\begin{minipage}{00.85\textwidth}
\par\vspace*{3ex}\noindent
{\bf Summary.~}
A rigorous empirical   comparison of two stochastic solvers is  important when
one of the solvers is a prototype of a new algorithm such as multiwalk (MWA).
When searching for global minima in $\mathbb{R}^p$, the key data structures of MWA include:
$\bm{p}$ rulers with each ruler assigned $\bm{m}$ marks and a set of $\bm{p}$ {\em neighborhood matrices} 
of size up to $\bm{m * (m-2)}$, where each entry represents absolute values of 
{\em pairwise differences} between $\bm{m}$ marks. Before taking the next step,
a controller links the tableau of neighborhood matrices and computes new and improved positions 
for each of the $\bm{m}$ marks. 
The number of columns in each neighborhood matrix is denoted as 
the {\em neighborhood radius} $\bm{r_n <= m-2}$.
Any variant of the  DEA (differential evolution algorithm) has
{\em an effective population neighborhood} of radius not larger than 1.
Uncensored first-passage-time performance experiments that vary the neighborhood radius of a MW-solver 
can thus be readily compared to existing variants of DE-solvers.
\par
~~This paper considers seven test cases of increasing complexity and demonstrates, under 
uncensored first-passage-time performance experiments:  
(1) significant variability in convergence rate for seven DE-based solver configurations, and
(2) consistent, monotonic, and significantly faster rate of convergence for the MW-solver prototype 
as we increase the neighborhood radius from 4 to its maximum value.
\end{minipage}
\end{center}

\begin{center}
{\footnotesize
\vspace*{3ex}
\begin{tabular}{p{0.8\textwidth}}
~ \\
\hline
\multicolumn{1}{|p{0.8\textwidth}|}{
NOTE: Unlike the original IEEE publication, this reprint has been typeset in \LaTeX~ using the vanilla 
document class `article'. This class generates more pages when compared to the 6-page limit of the IEEE original. Please cite the paper as follows:
{\em Franc Brglez. On Uncensored Mean First-Passage-Time Performance Experiments with Multiwalk in $\mathbb{R}^p$: a New Stochastic Optimization Algorithm. Invited talk, IEEE Proc. 7th Int. Conf. on Reliability, InfoCom Technologies and Optimization (ICRITO'2018); Aug. 29--31, 2018, Amity University, Noida, India, 2018.  
For the 6-page download, see
\url{https://people.engr.ncsu.edu/brglez/publications/OPUS2-2018-mwR-ICRITO-Brglez.pdf}
}
\par\vspace*{2ex} 
For on-going work and additional context, see
\par\vspace*{1ex} 
On Uncensored Global Stochastic Optimization in $\mathbb{R}^p$/$\mathbb{D}^p$ 
and Multi-Walk Algorithms under Problem-Specific Tableau Formulations
\par
\url{https://people.engr.ncsu.edu/brglez/publications/OPUS2-2018-mwRD-Brglez-talk.pdf}
 
\par\vspace*{1ex} 
Throwing Darts and Needles under Four Configurations: the Uncensored Mean First-Passage-Time of Hitting the $k$ Decimal Digits Value of $\pi$
\par
\url{https://people.engr.ncsu.edu/brglez/publications/OPUS2-2018-pi-tufte-Brglez.pdf}
}
\\
\hline
\end{tabular}
}
\end{center}
 \newpage

\vspace*{-4ex}
\begin{center}
{\large\bf 
On Uncensored Mean First-Passage-Time Performance Experiments \\[0.5ex]
with Multiwalk in $\mathbb{R}^p$: a New Stochastic Optimization Algorithm
} 
\par\vspace*{2ex}
Franc Brglez\\   
Computer Science, NC State University\\ 
Raleigh, NC 27695, USA\\
\href{mailto:brglez@ncsu.edu}{brglez@ncsu.edu}  
\vspace*{2ex}
\end{center}

\begin{multicols}{2} 
\noindent
{\bf Abstract.~}
A rigorous empirical   comparison of two stochastic solvers is  important when
one of the solvers is a prototype of a new algorithm such as multiwalk (MWA).
When searching for global minima in $\mathbb{R}^p$, the key data structures of MWA include:
$\bm{p}$ rulers with each ruler assigned $\bm{m}$ marks and a set of $\bm{p}$ {\em neighborhood matrices} 
of size up to $\bm{m * (m-2)}$, where each entry represents absolute values of 
{\em pairwise differences} between $\bm{m}$ marks. Before taking the next step,
a controller links the tableau of neighborhood matrices and computes new and improved positions 
for each of the $\bm{m}$ marks. 
The number of columns in each neighborhood matrix is denoted as 
the {\em neighborhood radius} $\bm{r_n <= m-2}$.
Any variant of the  DEA (differential evolution algorithm) has
{\em an effective population neighborhood} of radius not larger than 1.
Uncensored first-passage-time performance experiments that vary the neighborhood radius of a MW-solver 
can thus be readily compared to existing variants of DE-solvers.
\par
~~This paper considers seven test cases of increasing complexity and demonstrates, under 
uncensored first-passage-time performance experiments:  
(1) significant variability in convergence rate for seven DE-based solver configurations, and
(2) consistent, monotonic, and significantly faster rate of convergence for the MW-solver prototype 
as we increase the neighborhood radius from 4 to its maximum value.
 
\section{Introduction} 
\label{sec_introduction}
\noindent
A rigorous empirical  performance comparison of two stochastic solvers is of particular importance when
one of the solvers is new and under investigation for potential improvements.
The book on
{\em First-Passage Processes}~~\cite{OPUS-fpt-2001-CambridgeUP-Radner}
explains that 
\begin{quote}
... first passage underlies many stochastic processes in which the event, such as a dinner date, a chemical reaction, 
the firing of a neutron, or the triggering of a stock option relies on 
a variable reaching a specified value {\em  for the first time} ...
\end{quote}
\noindent
In the context of two stochastic solvers, the variable we monitor for reaching a specified value
is the {\em target value} of the objective function. 
Typically, the target value is also the {\em best-known-value} ({\em bkv}) 
since the {\em optimum value} may not been proven.
If the target value is not an integer, its value 
is specified by the total number of digits it contains before and after the decimal point.
We define the first-passage-time ({\em fpt}) stopping criterion for any solver as
the {\em stopping time} of the solver  when it returns
the target value {\em  for the first time}. 
We say that a solver run is {\em censored} if it stops due to a timeout limit before reaching the target value.
In this paper,
we compare the performance of two stochastic solvers by repeating the experiment 
with at least 100 random seeds and evaluate the answer to this question: 
{\em ``what is the uncensored mean time for each solver to reach the same target value?''}
We say that the comparison is reliable if at least one of the solvers has 0 censored runs.

\begin{figure*}[t!]
\vspace*{-2ex}
\centering

\includegraphics[width=0.49\textwidth]{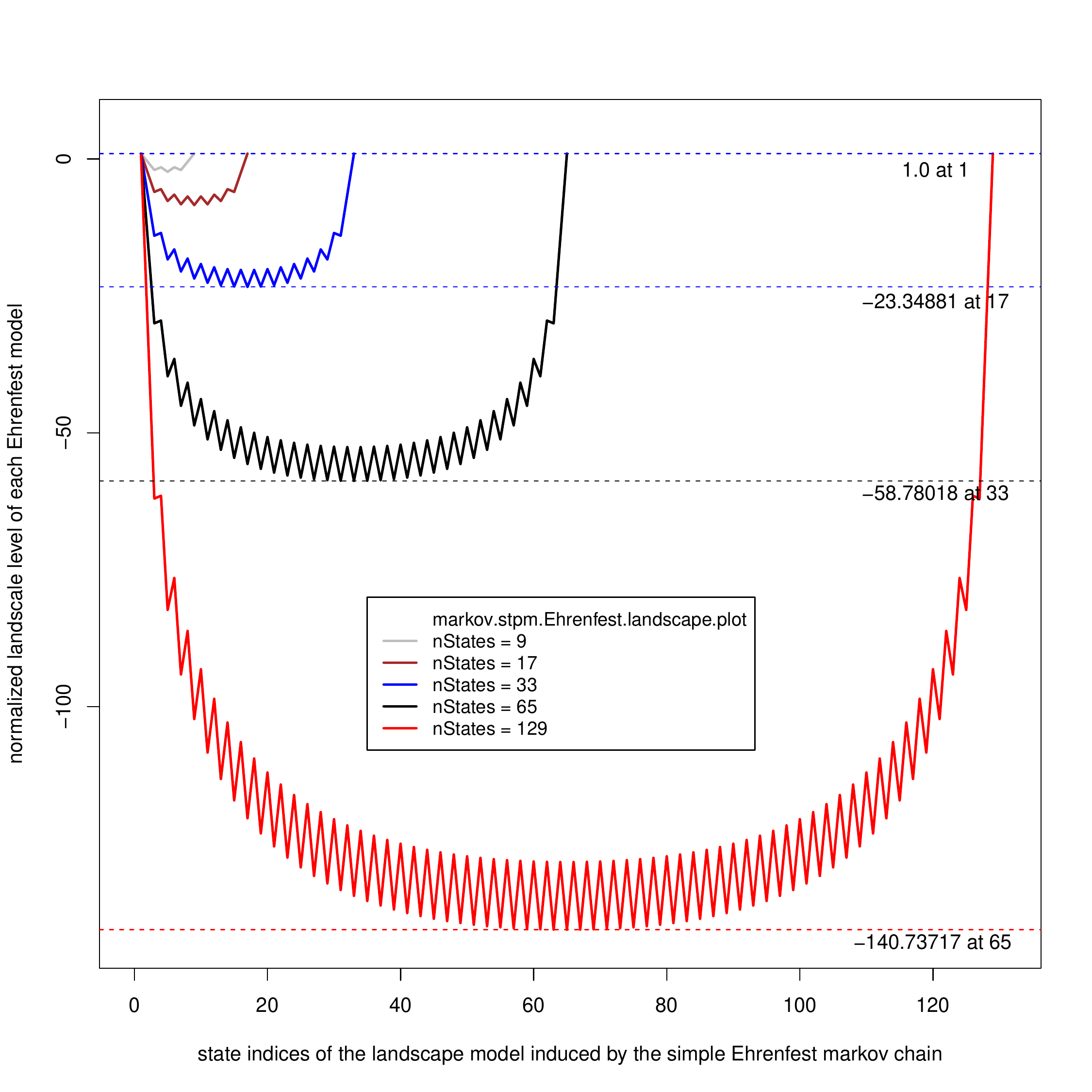}
\includegraphics[width=0.49\textwidth]{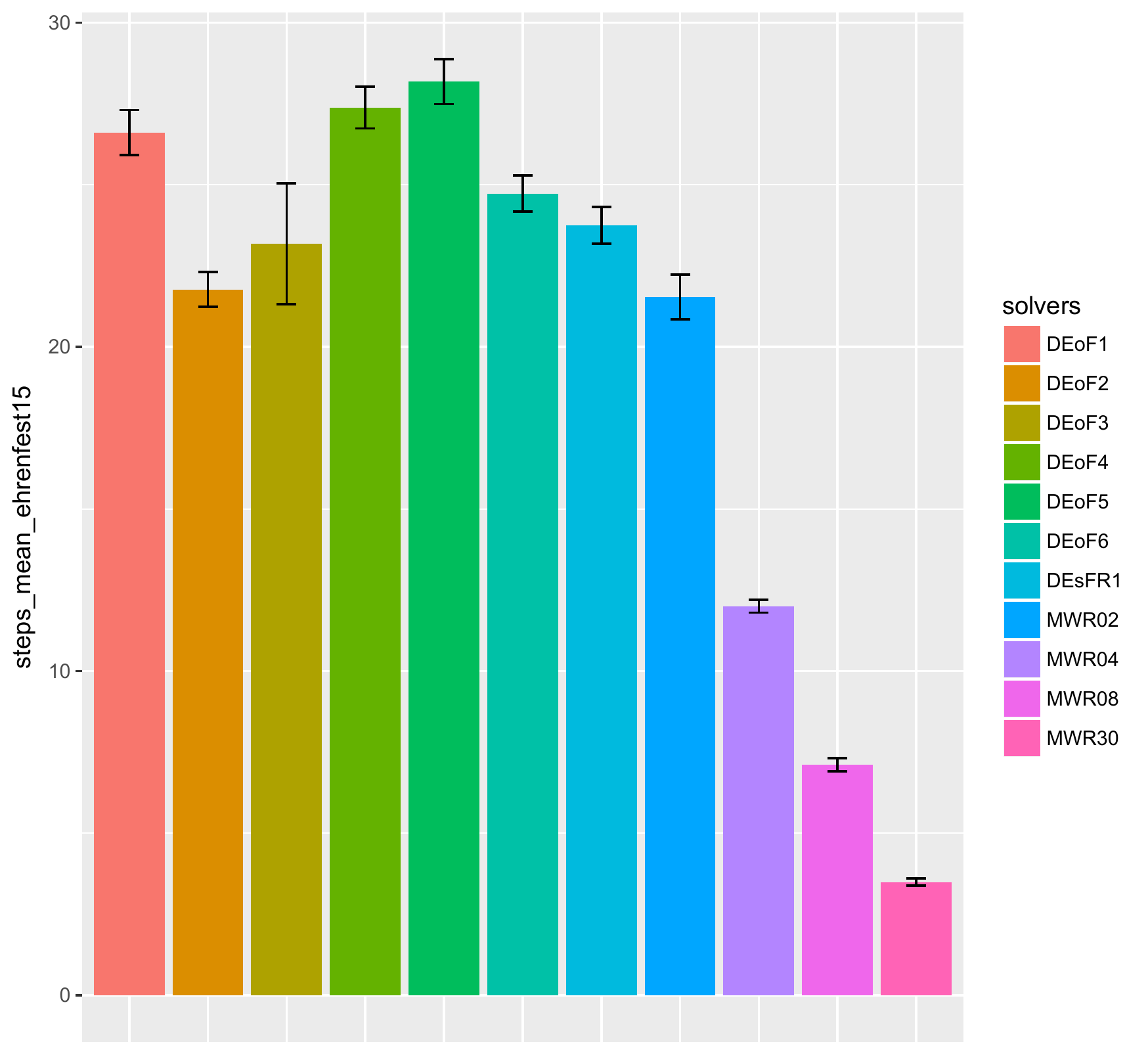}

\caption{
Summary of first-passage-time experiments  
with the function {\tt ehrefest} (a model representing $2^{15} + 1 = 32769$ states)
and eleven solvers. The first seven solvers implement documented variants of 
the well-known DE algorithm.
}
\label{fg_mw_barplot_ehrenfest}
\vspace*{2ex}
\end{figure*}

In contrast, typical computational experiments to rank the performance
stochastic optimization solvers are based on a much simpler approach: take $S$ solvers, $P$ problem instances, 
$N$ random seeds, run each solver under the stopping criterion of
{\em a fixed runtime limit}. Then, for each solver,
tabulate distances from best-known-values
and related statistics. In~\cite{OPUS-optim-2014-JSS-Mullen}, 
most insights are revealed in Figure 2 which tallies successes with 18 solvers over all 100 runs 
for each of 48 objective functions. Here is a verbatim quote: 
``A `success' was defined as a solution less than 0.005 more than the minimum of 
the objective function between the default bounds.''
In other words, {\em bkv} $\le$ {\em a solution value} $<$ {\em bkv} + 0.005.
Our  experiments with this stopping criterion show  that solver rankings become increasingly unreliable
as the percentage of censored results increases. 
This observation is also supported with arguments by statisticians~\cite{OPUS-statistics-1995-MathStat-Sen-censoring}.
Under criteria defined in Figure~2, the percentage of censored results ranges  from
{\tt (4800 - 3800)/4800 = 21\%} ~to {\tt (4800 - 1200)/4800 = 75\%}. 
If the error tolerance that defines a `success'  is reduced from 0.005 to 0.0005, 
the percentage of censored results in Figure 2 is most likely to increase rapidly towards 100\%.

In this paper,
the concept of first-passage-time and runtime limit is  measured in units that
are platform-independent: on a granular scale we count the number of objective function
evaluations ({\em probes}). On a higher level, we report the rate of solver convergence towards the target value
by counting the number of iterations or steps. We replace the value of runtime limit with the
value of {\em iterations/steps limit}.

\section{Background and Motivation}
\label{sec_background}
\noindent
In a seminal paper Kac explains the  Ehrenfest model of diffusion   
with an $s*s$ state-transition probability matrix 
and makes a connection to random walks 
on graphs~\cite{OPUS-markov_ehrenfest-1947-AMM-Kac-Random_walk-Brownian_Motion}.
As part of an on-going research to be reported elsewhere, we have transformed
this  matrix to an objective function
{\tt ehrenfest(x)} defined on the set of integers in the range $[1, s]$.
See Figure~\ref{fg_mw_barplot_ehrenfest} for plot of function values for $s \in (9, 17, 33, 65, 129)$.
The adjacent bargraph represents a template that summarizes 
a statistical experiment with sampleSize = 100, reporting the mean values of
steps returned by each of 11 solver configurations upon finding
the minimum value solution for the function
{\tt ehrenfest15(x)}. All parameters relevant to results in this bargraph are summarized 
in the table below:
\par\vspace*{-1ex}
\begin{Verbatim}[frame=lines, fontsize=\scriptsize, xleftmargin=0mm]
      OFname = ehrenfest15       solvers are configured for
    nStates  = 2^15 + 1          first-passage-time stopping 
   coordBest = 16384, 16386      DEoF1,  DEoptim, strategy=1
 valueTarget = -78544.9529       DEoF2,  DEoptim, strategy=2
digitsTarget = 9                 DEoF3,  DEoptim, strategy=3
       OFtol = 5e-04             DEoF4,  DEoptim, strategy=4
 rulerMarks  = 32                DEoF5,  DEoptim, strategy=5
     agentId = 1,2,..,32         DEoF6,  DEoptim, strategy=6
      dither = 0.01          DEsFR1, simpleDE, restarts, r=1
neighbRadius = 2,4,8,30       MWRxx, multiwalk,  xx=2,4,8,30
\end{Verbatim}
%
We briefly explain the solvers and the most important names and values of variables in this table.
The first six solvers, {\tt DEoF1} to {\tt DEoF6} represent six configurations of the same solver 
{\tt DEoptim}, readily accessible as an R-package~\cite{OPUS-R-2011-JSS-Mullen}. 
A very useful property of {\tt DEoptim} is that it accepts a user-defined entry for {\em valueTarget},
which then allows for implementation of the first-passage-time stopping criterion. Importantly, we must  pass
the variable {\em targetDigits} to the objective function so that the value returned by the objective
function is 'quantized' with the R-command 'signif'.
For example, {\tt signif(1234.5789 - 0.0004999, 9)} returns 1234.5784 in R-shell.

The solver {\tt DEsFR1} is our extension of 'simpleDE' code  supplied with the R-package 'adagio'~\cite{R-package-2018-adagio-Borchers}. The code has been extended to support both the first-passage termination criterion
as well as {\em restarts}, matching the capabilities of solver {\tt HWR}.
The most important data structures in {\tt HWR} are the {\em ruler}, the {\em maximum ruler neighborhood},
and the {\em neighborhood radius}. The number of {\em marks} in the ruler is equivalent to the {\em size of population} in
DE-based solvers. 
For details, see Section~\ref{sec_MWA}.

%

%

\begin{figure*}[t!]
\centering

\hspace*{-0em}
\begin{minipage}{0.46\textwidth}
In mathematics, a ruler is a set of marks in $\mathbb{R}$ or  
$\mathbb{Z}$. A function such as {\tt ehrenfest4}  in
Figure~\ref{fg_mw_barplot_ehrenfest} is defined 
on the range $[1, 2^4 +1=17]$. In this example we choose
a ruler with 6 marks. For illustration
we choose all initial marks as integers (1, 2, 4, 10, 12, 17),
relating them as close as possible to the familiar optimal 
Golomb ruler with 6 marks~\cite{OPUS-ogr-2018-WWW-Wikipedia}.
A very important data structure for the MW algorithm
is the {\em maximum ruler neighborhood},
illustrated below: we create it by  
first constructing the 
adjacent {\em ruler difference matrix}
with all pairs of marks in the ruler.

\begin{Verbatim}[frame=lines, fontsize=\footnotesize,
xleftmargin=0mm]
ruler = (1, 2, 4, 10, 12, 17)
   ruler difference matrix     ruler neighborhood
   [1] [2] [3] [4] [5] [6]       [1] [2] [3] [4]
1  NA   2   4  10  12  17         2   4  10  12
2   4  NA   3   9  11  16         3   9  11  16
3   4   3  NA   7   9  14         3   7   9  14
4  10   9   7  NA   3   8         9   7   3   8
5  12  11   9   3  NA   6        11   9   3   6
6  17  16  14   8   6  NA        16  14   8   6
\end{Verbatim}     
{\footnotesize
  
}
\end{minipage}
\begin{minipage}{0.03\textwidth}
~~~~ 
\end{minipage}
\begin{minipage}{0.49\textwidth}
\centering 
\vspace{-2ex}
\includegraphics[width=0.99\textwidth]{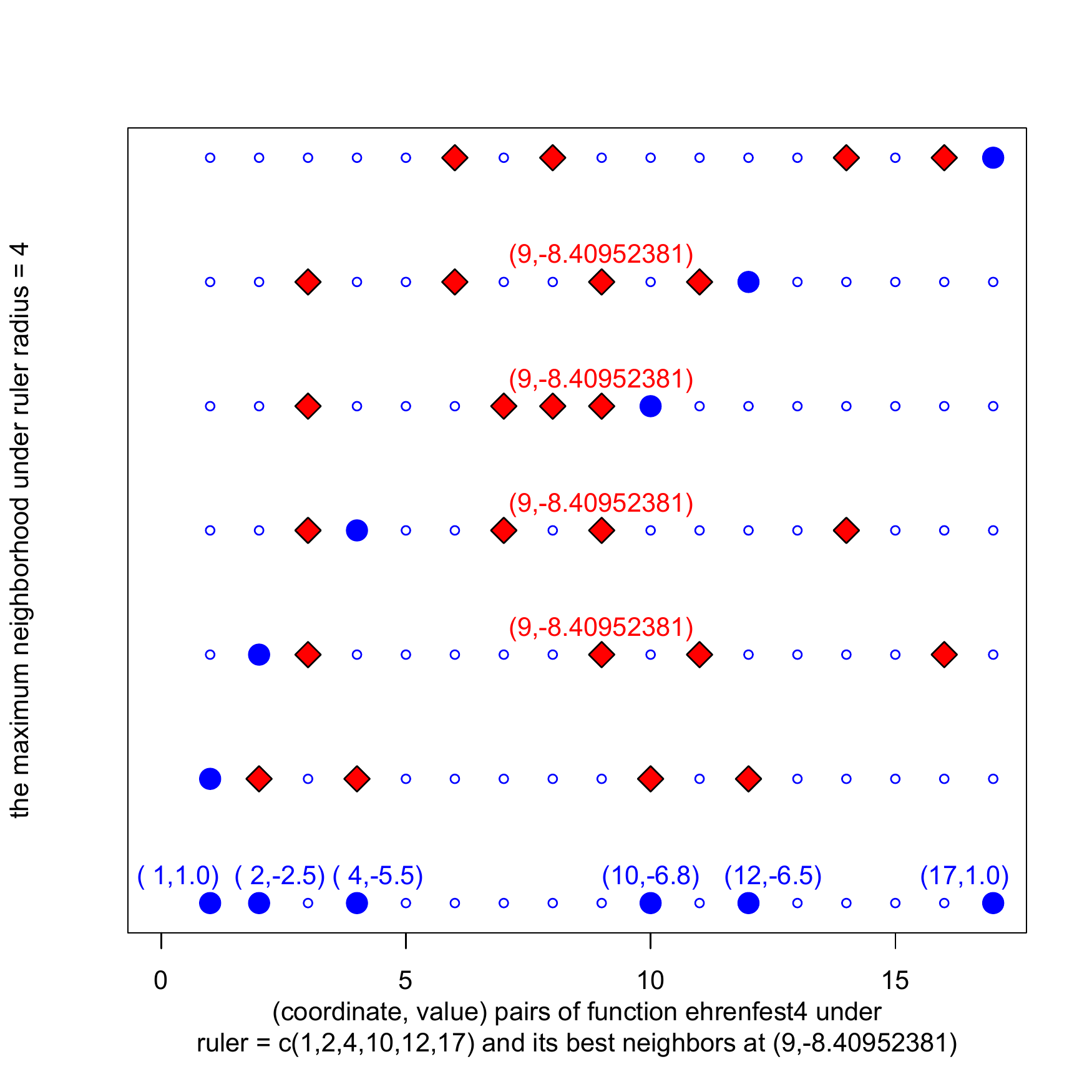}
\end{minipage}
\caption{
A ruler with 6 marks and its maximum neighborhood under ruler radius of 4.
For the initial ruler = (1,2,4,10,12,17), the MW solver finds
the minimum value of function {\tt ehrenfest4} in a single step.
With population size of 6, DE solvers cannot solve this instance
in thousands of steps. However, {\em very rarely}, a DE solver 
has found a solution under 500 steps!}
\label{fg_mw_ruler_neighborhood}
\vspace*{-0ex}
\end{figure*}

By default, {\em valueTarget, digitsTarget} represent the best-known-value of the objective function,
expressed with 9 digits. In the spirit of~\cite{OPUS-opt-2004-SIAM-Bornemann}, 
we maintain digitsTarget = 9 for each valueTarget associated with each objective function and with each solver.



The consistent performance of the four multiwalk solver  configurations {\tt HWRxx} is a great motivator
to review its details in the next section. As we increase the neighborhood radius from 2 to the maximum of 30,
the mean number of steps 
reduces from 21.5 to 3.5! 

For DE-based solvers, 
the best best mean value of 21.77 steps is returned by solver {\tt DEoF2} under the strategy=2 configuration.
The increase in standard error observed for solver {\tt DEoF3} is due to a single run that has been censored at 200 steps.
The  next best mean value of 23.75 steps is returned by solver {\tt DEsF1}, 
our extension of 'simpleDE' code  supplied with the R-package 'adagio'~\cite{R-package-2018-adagio-Borchers}.
The bargraph in Figure~\ref{fg_mw_barplot_ehrenfest} is a template for
the harder test functions introduced in Figure~\ref{fg_mw_barplot_hard}.
%

\section{The Multi-Walk Algorithm (MWA)}
\label{sec_MWA}
\noindent
To outline the intuition that underlies the multi-walk  algorithm 
without loss of generality, we use
a simple example of  search for the minimum of the function {\tt ehrenfest(x)} in Figure~\ref{fg_mw_barplot_ehrenfest}.
The function {\tt ehrenfest4(x)} is defined on the range $[1, 17]$: we select randomly 4 points in this range,
say 4,12,10,2. The choice of integers is for simplicity only. 
By combining the two end points from the range
and the four random points 
into an ordered arrangement of $m=6$~{\em marks},
we construct the {\em ruler}:
\[ {\rm ruler} = (1,2,4,10,12,17) \]
Next, we consider a complete graph with $m=6$ vertices and $m*(m-1)=30$ edges, 
where marks serve as coordinates for each vertex. 
We define weight of each edge  as the absolute value of differences between each pair of marks.
The resulting structure is called the {\em ruler difference matrix}, shown in 
Figure~\ref{fg_mw_ruler_neighborhood}. Creating this matrix is only an intermediate step,
what we need is the {\em ruler neighborhood matrix} next to it: it has $m=6$ rows and $m-2$ columns.
The number of columns in each neighborhood matrix is denoted as {\em neighborhood radius} $r_n <= m-2$.
The red marks in the adjacent plot represent coordinate positions of
difference in the ruler neighborhood matrix. Moreover, the ruler coordinates at the bottom
of this plot are presented as (coordinate,value) pairs where each value is computed by evaluating
the function {\tt ehrenfest4(x)}. Since, for this function,
valueTarget=-8.40952381, the pair (9,-8.40952381) 
is the solution found by this search 
already on step=1.



\begin{figure*}[t!]
\hspace*{-2.1em}
\begin{minipage}{0.51\textwidth}
\centering
\begin{Verbatim}[frame=lines, fontsize=\scriptsize,numbers=left,
numbersep=3pt,firstline=1,xleftmargin=9mm]
MWR = function(OFname, pLB, pUB, rulerMarks, valueTarget, 
      digitsTarget, stepsLmt, dither, seed, ...) 
{ 
  fun = match.fun(OFname) 
  m   = rulerMarks  ;# number of ruler marks
  p   = length(pLB) ;# number of rulers
  w   = OF$neighbRadius
  neighbSize = m*p*w ; neighbRadius = w
  L = matrix(rep(pLB,each=m), nrow=m, ncol= p) 
  U = matrix(rep(pUB,each=m), nrow=m, ncol= p) 
  
  set.seed(seed) ;# initialize RNG
  R = matrix(runif(m*p), nrow=m, ncol=p)  
  R = L + R * (U - L) ; R[1,] = pLB ; R[m,] = pUB
  F = NULL   
  for (i in 1:m) {F = c(F , fun(R[i,]))}
  Rnext     = R   ; probes = OF$probes 
  valueBest = Inf ; steps  = 0 
   
  while (TRUE) { ;# count steps
    steps = steps + 1
    if (neighbRadius == m-2) {
      neighbors = ruler_neighborhood_eval( 
                  R,p,OFname,dither)  
    } else {
      neighbors = ruler_neighborhood_eval_rand(
                  R,p,neighbRadius,OFname,dither)  
    }
    probes = OF$probes 
    for (i in 1:m) {
      ci  = neighbors$Rnext[i,]
      fi  = neighbors$Fnext[i]  
      if (fi < F[i]) {
        Rnext[i,] = ci ; F[i]  = fi
      }
      if (fi < valueBest) {
        coordBest = ci 
        valueBest = signif(fi, digitsTarget)   
      }
    }
    R = Rnext 
    if (valueBest == valueTarget) {
      isCensored  = FALSE ; break
    }
    if (steps     >= stepsLmt)    {
      isCensored = TRUE ; break
    }
  } # end steps
  agentId = which.min(F) 
  return(c(coordBest,valueBest,agentId,steps, ...))
} 
\end{Verbatim}
\vspace*{-4.5ex}
\begin{Verbatim}[frame=lines, fontsize=\scriptsize,numbers=left,
numbersep=3pt,firstline=1,xleftmargin=9mm]
  In our 'DEfpt/DEfptR' prototypes, 'ci, fi' are  
  based on expressions from original code 'simpleDE' 
  below, with 'isConfined = TRUE' and 'rDE = 1.0':
    ...
    for (i in 1:m) {
      ii = sample(1:m, 3)
      ci = R[ii[1],] + rDE*(R[ii[2],] - R[ii[3],])  
      if (isConfined) {
        if (any(ci < pLB) || any(ci > pUB)) {
          ci = pLB + runif(p)*(pUB - pLB)  
        }
      }
      fi = fun(ci) 
      ...
    ...
  ...
  \end{Verbatim}
\end{minipage}
\begin{minipage}{0.51\textwidth}
\centering
\vspace*{-1ex}
\begin{Verbatim}[frame=lines, fontsize=\scriptsize,numbers=left,
numbersep=3pt,firstline=1,xleftmargin=9mm]
MWR = function(OFname, pLB, pUB, rulerMarks, valueTarget, 
      digitsTarget, stepsLmt, dither, seed, plateauLmt, ...) 
{ 
  fun = match.fun(OFname) 
  m   = rulerMarks  ;# number of ruler marks
  p   = length(pLB) ;# number of rulers
  w   = OF$neighbRadius
  neighbSize = m*p*w ; neighbRadius = w
  L = matrix(rep(pLB,each=m), nrow=m, ncol=p) 
  U = matrix(rep(pUB,each=m), nrow=m, ncol=p) 
  restarts = -1 ; steps = 0   
  
  while(TRUE) { ;# count restarts
    restarts = restarts + 1
    set.seed(seed) ;# initialize RNG
    R = matrix(runif(m*p), nrow=m, ncol=p)  
    R = L + R * (U - L)  ; R[1,]=pLB ; R[m,]=pUB 
    F = NULL ; for (i in 1:m) {F= c(F,fun(R[i,]))}
    iMin = which.min(F)
    errPrev = F[iMin] - valueTarget
    Rnext      = R   ; probes = OF$probes     
    valueBest = Inf  ; steps2 = 0 ; plateauCnt = 0
      
    while (TRUE) {  ;# count steps 
      steps2 = steps2 + 11
      if (neighbRadius == m-2) {
        neighbors = ruler_neighborhood_eval( 
                    R,p,OFname,dither)  
      } else {
        neighbors = ruler_neighborhood_eval_rand(
                    R,p,neighbRadius,OFname,dither)  
      }
      probes = OF$probes 
      for (i in 1:m) {
        ci  = neighbors$Rnext[i,]
        fi  = neighbors$Fnext[i]  
        if (fi < F[i]) {
          Rnext[i,] = ci ; F[i]  = fi
        }
        if (fi < valueBest) {
          coordBest = ci 
          valueBest = signif(fi, digitsTarget)   
        }
      }
      R = Rnext ; error = valueBest - valueTarget
      if (error == 0) {isCensored = FALSE ; break}
      if (error >= errPrev) {
        # no reduction of error 
        plateauCnt = plateauCnt + 1
      } else {
        # error reducing step; reset plateauCnt
        plateauCnt = 0 ; errPrev = error  
      }
      if (plateauCnt == plateauLmt) {
        isCensored = TRUE ; break ;# and restart
      }
      if (steps2 >= stepsLmt) {isCensored=TRUE ; break}
    } # end steps2
    steps = steps + steps2
    agentId = which.min(F) 
    if (!isCensored)       {break}
    if (steps >= stepsLmt) {isCensored=TRUE ; break}
    
    # new seed for the next restart
      seed = round(0.5 + 1e9*runif(1))
    } # end restarts
    return(c(coordBest,valueBest,agentId,steps, ...))
}
\end{Verbatim}
\end{minipage}
\vspace*{0ex}
\caption{
An {\em fpt} implementation of 
the multi-walk algorithm ({\em MWA} without/with restarts  ({\tt MW}/{\tt MWR})
and a snippet from {\tt DEfptR}. 
}
\label{fg_mw_stopping_templates}

\end{figure*}

\begin{figure*}[t!]
\vspace*{-0ex}
\centering

\hspace*{-0em}
\begin{minipage}{0.99\textwidth}
\centering 
\includegraphics[width=0.99\textwidth]{Figures/fg_mw_walks_}
\end{minipage}

\par\vspace*{-4ex}
\includegraphics[width=0.49\textwidth]{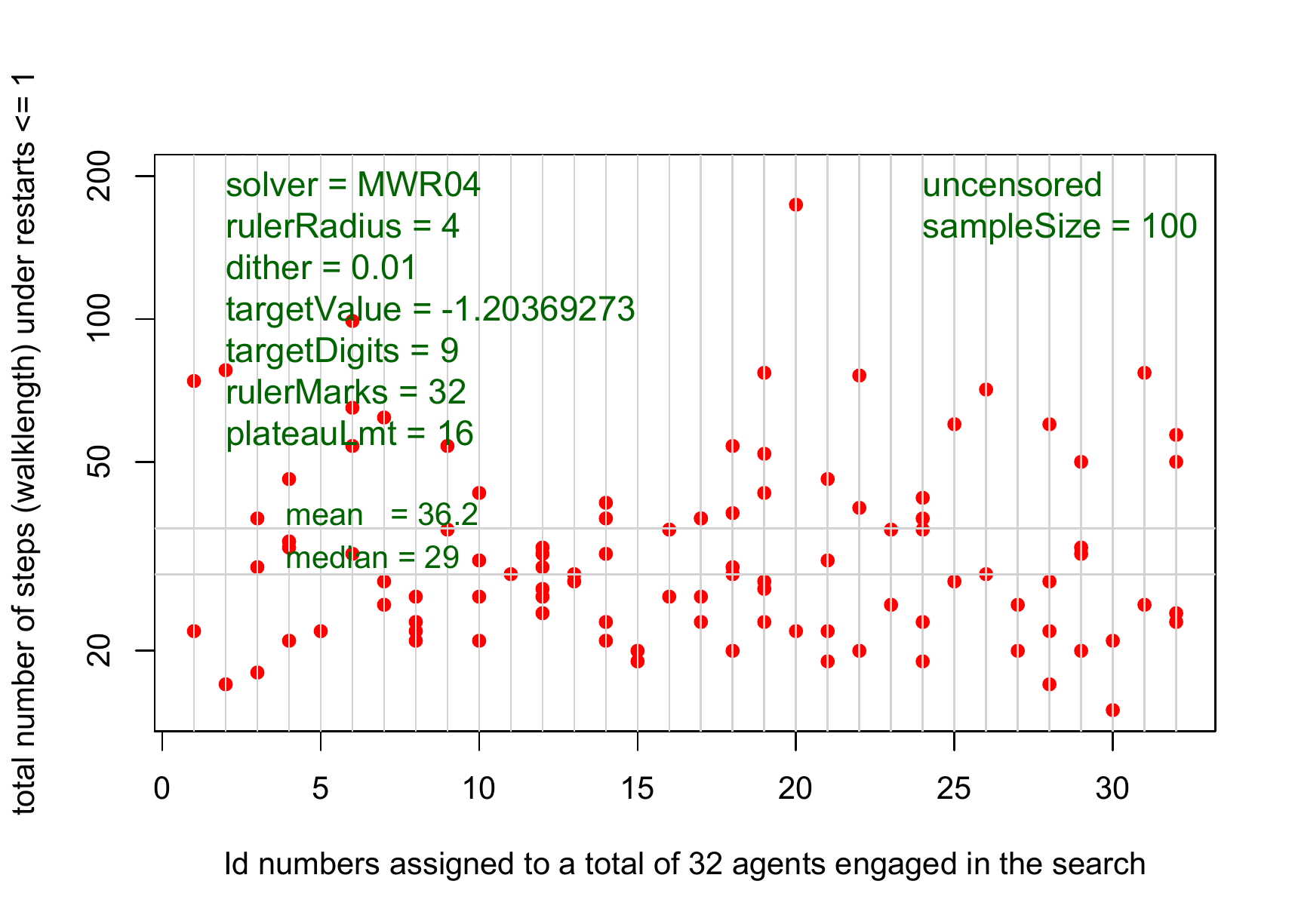}
\includegraphics[width=0.49\textwidth]{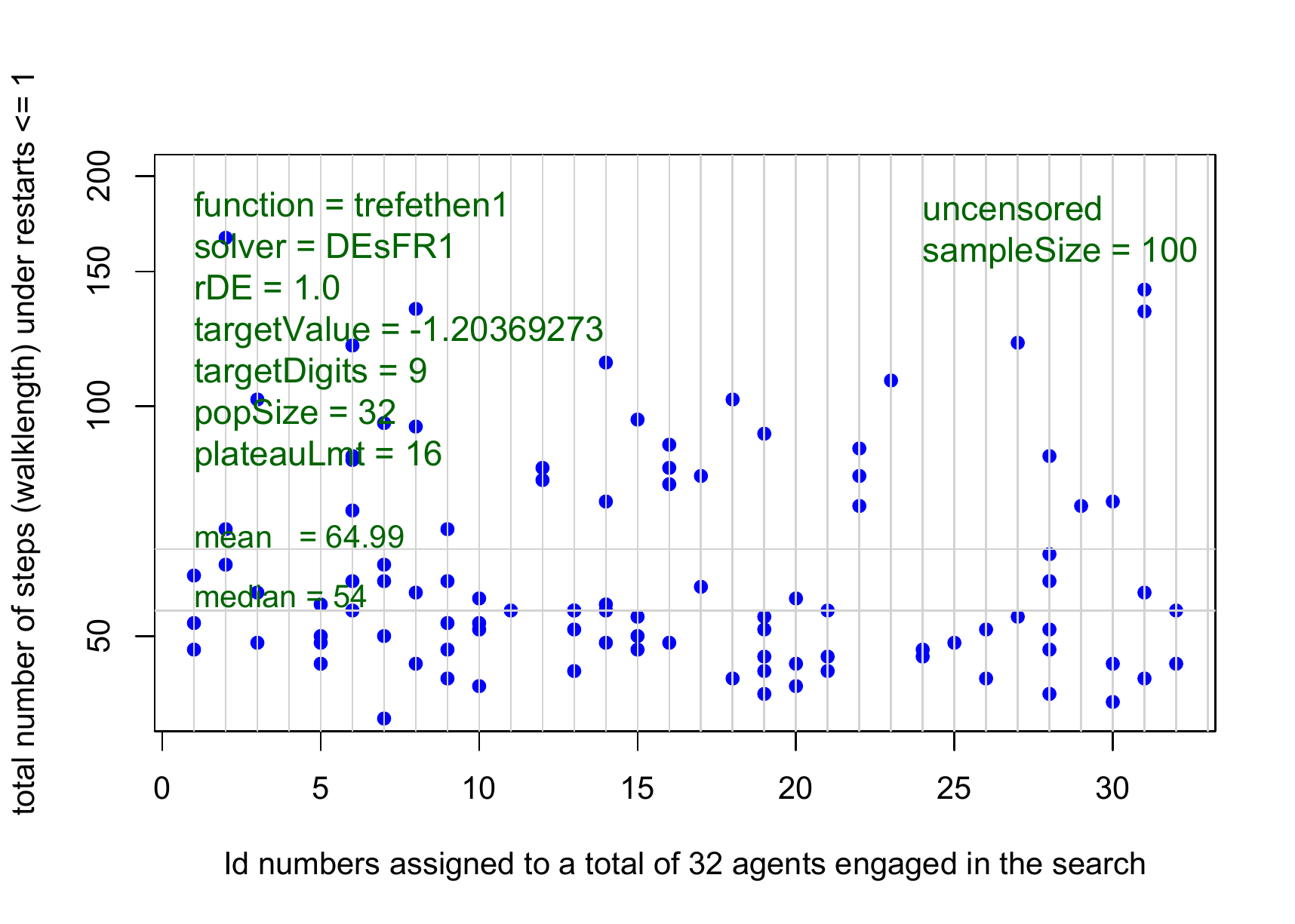}

\caption{
Walks controlled by two solvers, {\tt MWR04} and {\tt DEsFR1}.
Each ruler mark is assigned an {\em agentId}. 
Walks with dotted lines are controlled by the value of {\em plateauLmt}.
Solid line associates with walk and agentId that reaches the {\em valueTarget} first.  
}
\label{fg_mw_walks}
\vspace*{-0ex}
\end{figure*}

For steps $s = 0, 1, 2, \ldots$, the multi-walk can be formulated recursively:
\[({\cal R}^{s+1}, {\cal F}^{s+1}) = \Psi({\cal R}^s, {\cal F}^s, {\cal RN}^s) \] 
\hspace*{1.3em}${\cal R}$~~ ..... the matrix of ruler coordinates \\
\hspace*{1.3em}${\cal F}$~~ ..... the matrix of objective function values\\
\hspace*{1.3em}${\cal RN}$ .... the ruler neighborhood matrix \\
\hspace*{1.3em}$\Psi$~ ..... the R-based implementation in Figure~\ref{fg_mw_stopping_templates}

\vspace*{1ex}\noindent
The implementation of {\em MWA} in Figure~\ref{fg_mw_stopping_templates} has two main parts:
the left column implements {\tt MW} without the support for random restart, 
the column on the right implements {\tt MWR} which  supports a random restart.
At the bottom of the column on the right,
we show a snippet of the original code from~\cite{R-package-2018-adagio-Borchers}.
A statement {\tt agentId = which.min(F)} is contained in 
both {\tt MW} and {\tt MWR}.  This statement accesses the value of
not only {\em valueBest} of the objective function but also   {\em agentId}
as a number   from the range $[1, m]$;  a number  reported as the index of the mark that reaches
{\em valueTarget}.
See Figure~\ref{fg_mw_walks}.

There are traces of four walks in Figure~\ref{fg_mw_walks}. 
The MW-solver engaged a ruler with 32 marks and {\em a radius of 4}, i.e. only 4 neighbors (from the maximum of 30) are considered 
as candidates for the next step. Since agentId=32 is the first to reach the target value on step=59, the walk with solid line reports its
position for the full duration of the walk. We have a similar arrangement for DE-solver where agentId=7 is the first to reach the same
target value, but now on step=95. 

Results in Figures~\ref{fg_mw_ruler_neighborhood} and~\ref{fg_mw_walks}
support our intuition that underlies the multiwalk  algorithm. By associating the ruler-based coordinates
with differences of such coordinates creates the global neighborhood as the key to accelerating the convergence
of {\em MWA}.
The most significant observation we make about the experimental results in Figure~\ref{fg_mw_walks}
is this: a neighborhood with a radius of only 4 (from a maximum of 30) reduces the number of steps from the mean 
value of 64.99 for the DE-solver to 36.2 for the MW-solver. The best choices of parameters such as {\em dither} 
that adds a controlled amount of noise to each entry in the neighborhood matrix (default is at 1\% or less),
and the {\em tableauLmt }(default is the number of marks in each ruler) to control restarts, will be discussed elsewhere.

\section{Experiments and Compararisons}
\label{sec_experiments}
\begin{figure*}[!t]
\centering

\hspace*{0em}
\begin{minipage}{0.46\textwidth}
{\footnotesize
trefethen1, best DE-solver = DEsFR1: steps\_mean = 64.99\\
trefethen1, best MW-solver = MWR30:  steps\_mean = 13.33\\
trefethen1, best DE/MW  = 64.99/13.33 = 4.88\\
trefethen2, best DE-solver = DEsFR1: steps\_mean = 281.38\\
trefethen2, best MW-solver = MWR30:  steps\_mean = 86.36 \\
trefethen3, best DE/MW  = 281.38/86.36 = 3.25\\
trefethen3, best DE-solver = DEsFR1: steps\_mean = 532.15\\
trefethen3, best MW-solver = MWR30:  steps\_mean = 121.34\\
trefethen3, best DE/MW  = 532.15/121.34 = 4.39\\
{\color{red}{\sf most outlier bars are heavily censored!}}
}
\end{minipage}
\begin{minipage}{0.03\textwidth}
~~~~ 
\end{minipage}
\begin{minipage}{0.49\textwidth}
\hspace*{-2.2em}
\includegraphics[width=0.99\textwidth]{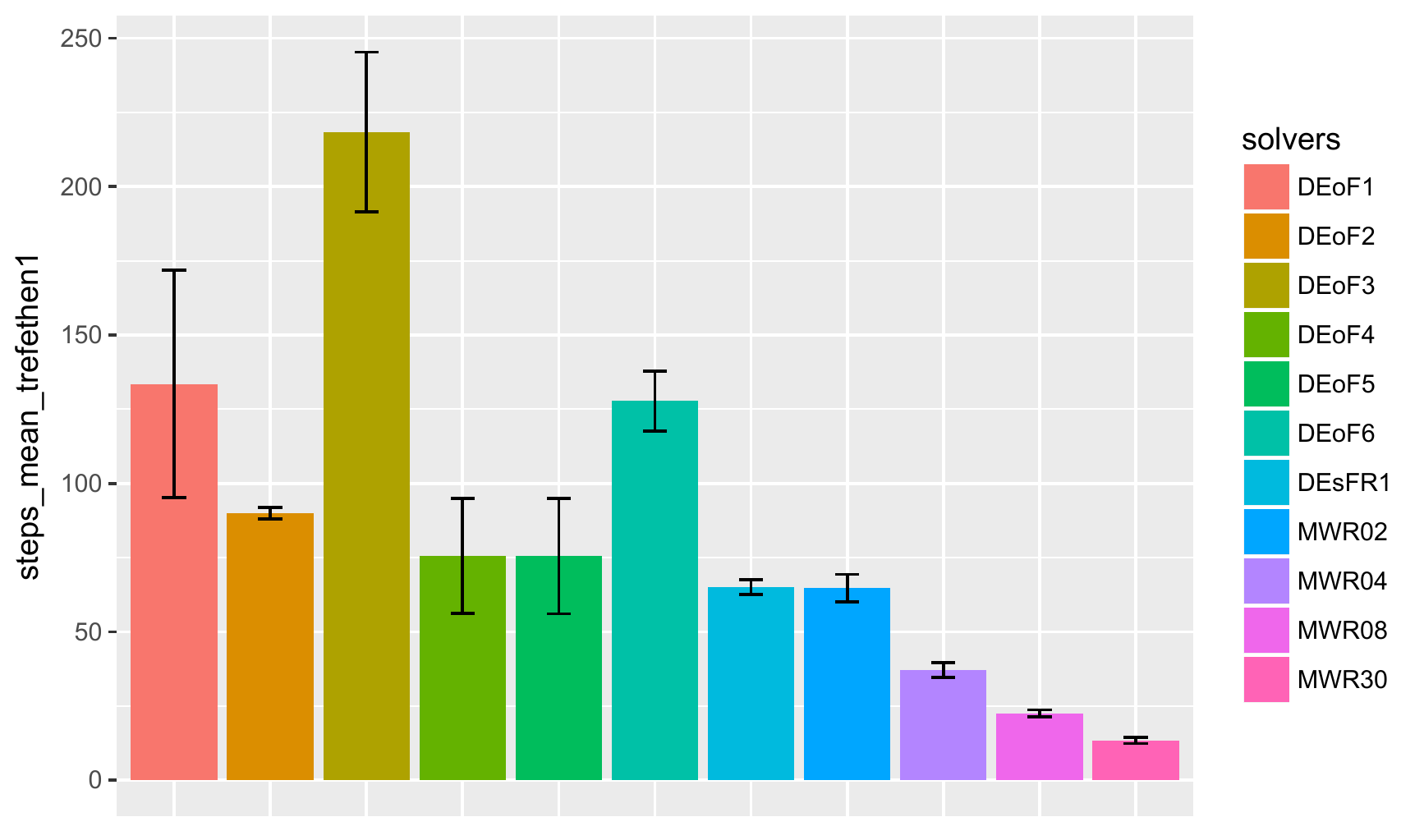}
\end{minipage}

\includegraphics[width=0.45\textwidth]{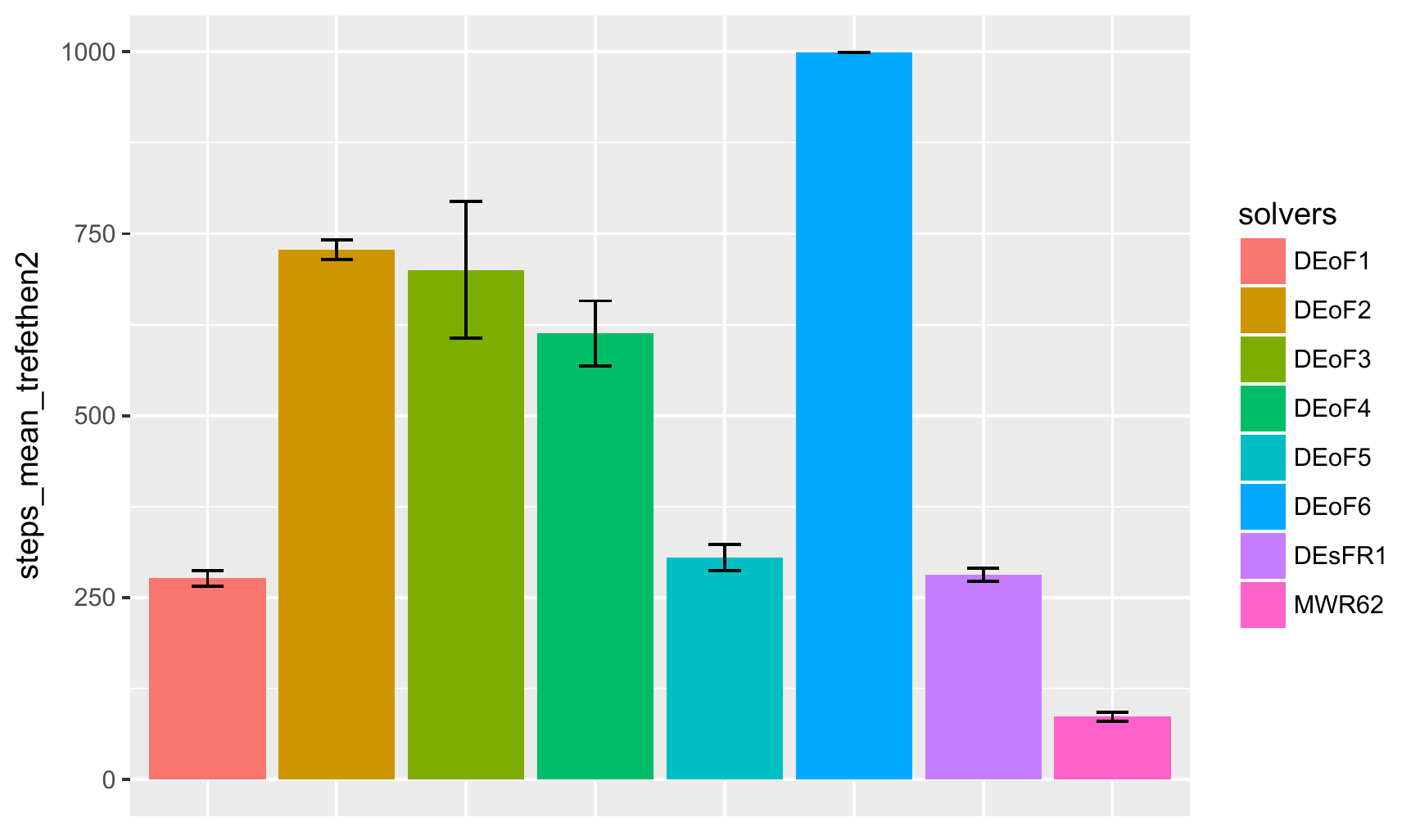}
\includegraphics[width=0.45\textwidth]{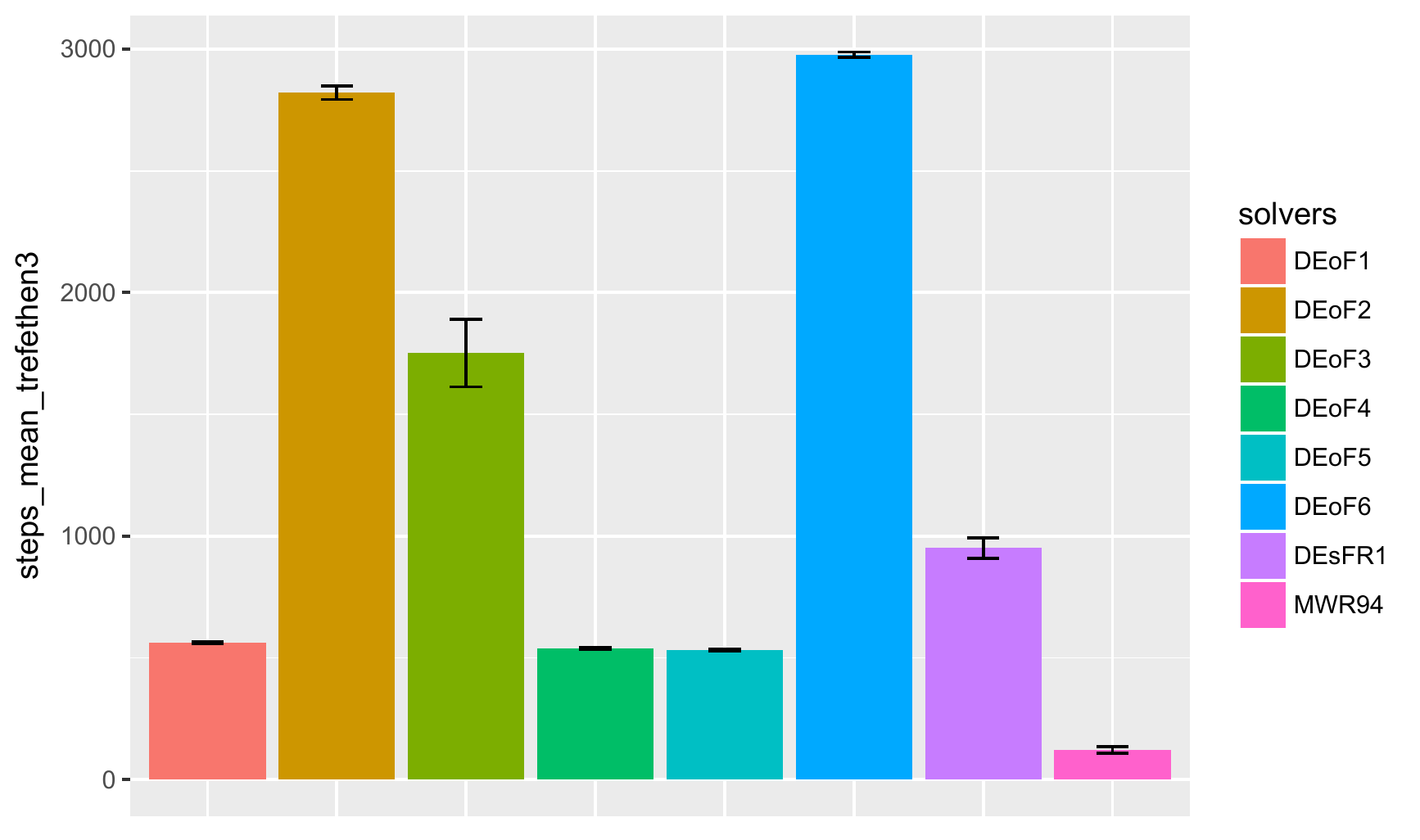}

\vspace*{4ex}
\hspace*{-0em}
\begin{minipage}{0.46\textwidth}
{\footnotesize
wild1, best DE-solver = DEoF5: steps\_mean =  68.47\\ 
wild1, best MW-solver = MWR30:  steps\_mean = 20.83\\
wild1, best DE/MW  = 68.47/20.83 = 4.15\\
wild2, best DE-solver = DEsFR1: steps\_mean = 188.49\\
wild2, best MW-solver = MWR30:  steps\_mean = 21.0 \\
wild2, best DE/MW  = 188.49/21.0 = 8.98\\
wild3, best DE-solver = DEsFR1: steps\_mean = 532.15\\
wild3, best MW-solver = MWR30:  steps\_mean = 17.87\\
wild3, best DE/MW     = 532.15/17.87        = 29.8\\
{\color{red}{\sf most outlier bars are heavily censored!}}
}
\end{minipage}
\begin{minipage}{0.03\textwidth}
~~~~ 
\end{minipage}
\begin{minipage}{0.49\textwidth}
\hspace*{-2.2em}
\includegraphics[width=0.99\textwidth]{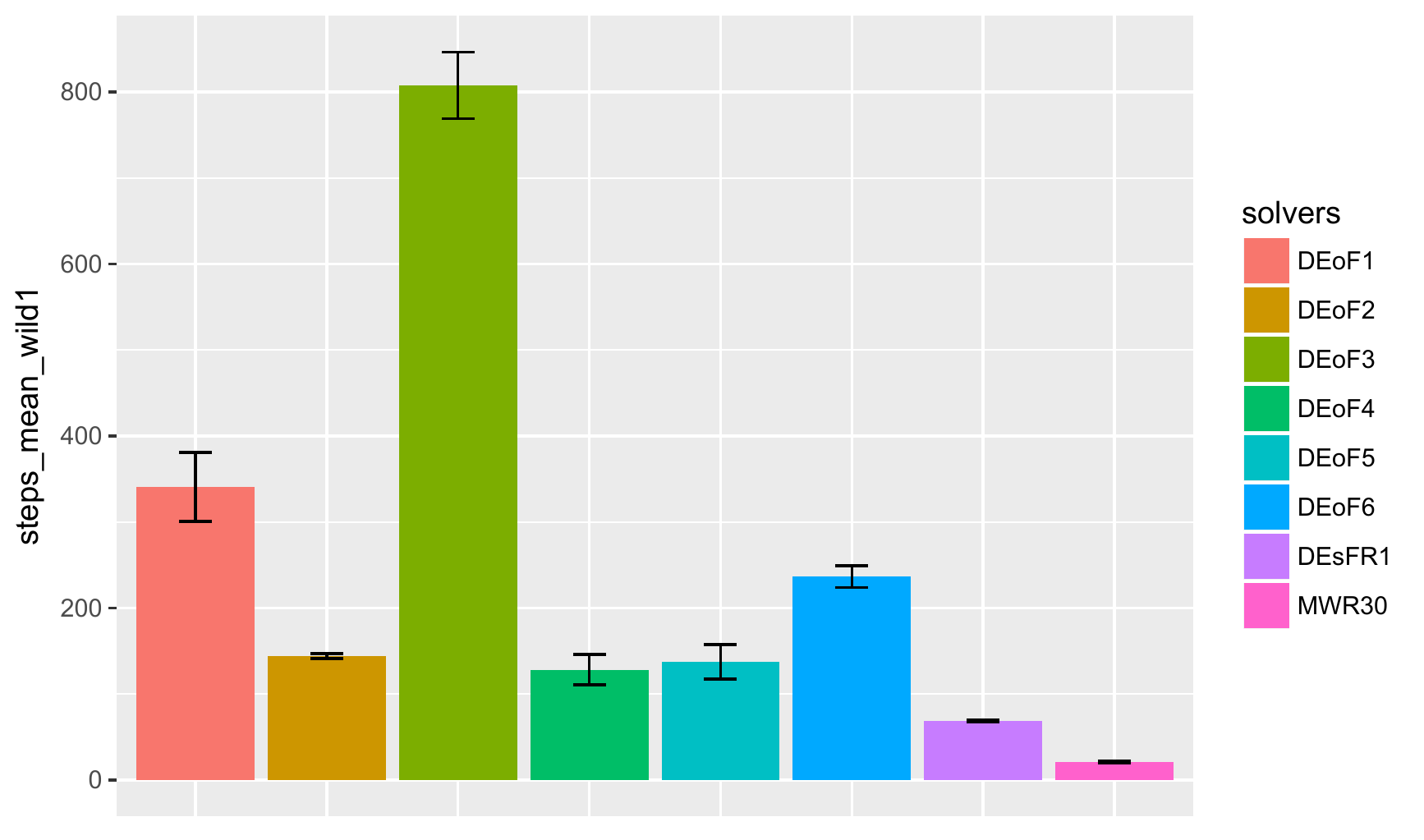}
\end{minipage}

\includegraphics[width=0.45\textwidth]{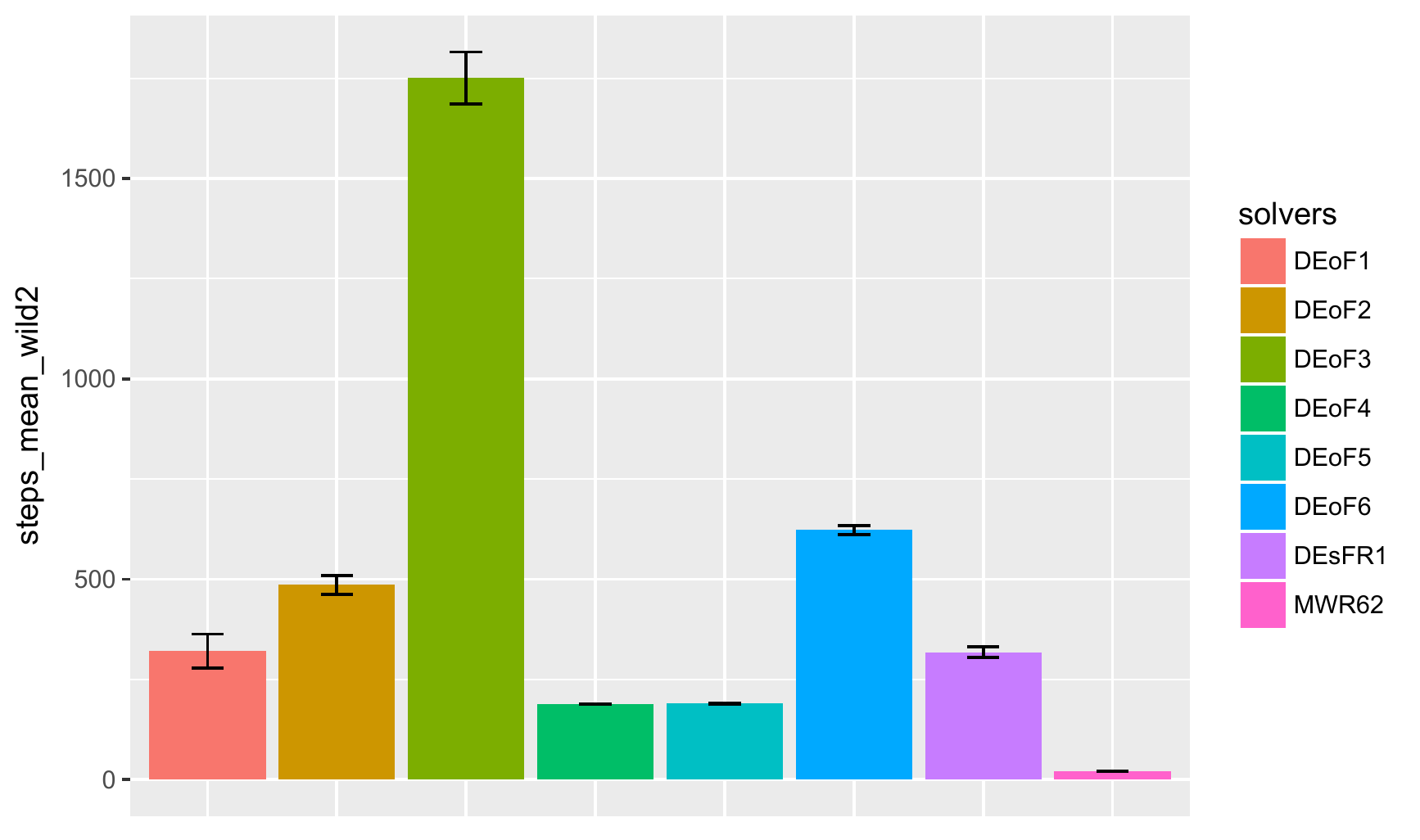}
\includegraphics[width=0.45\textwidth]{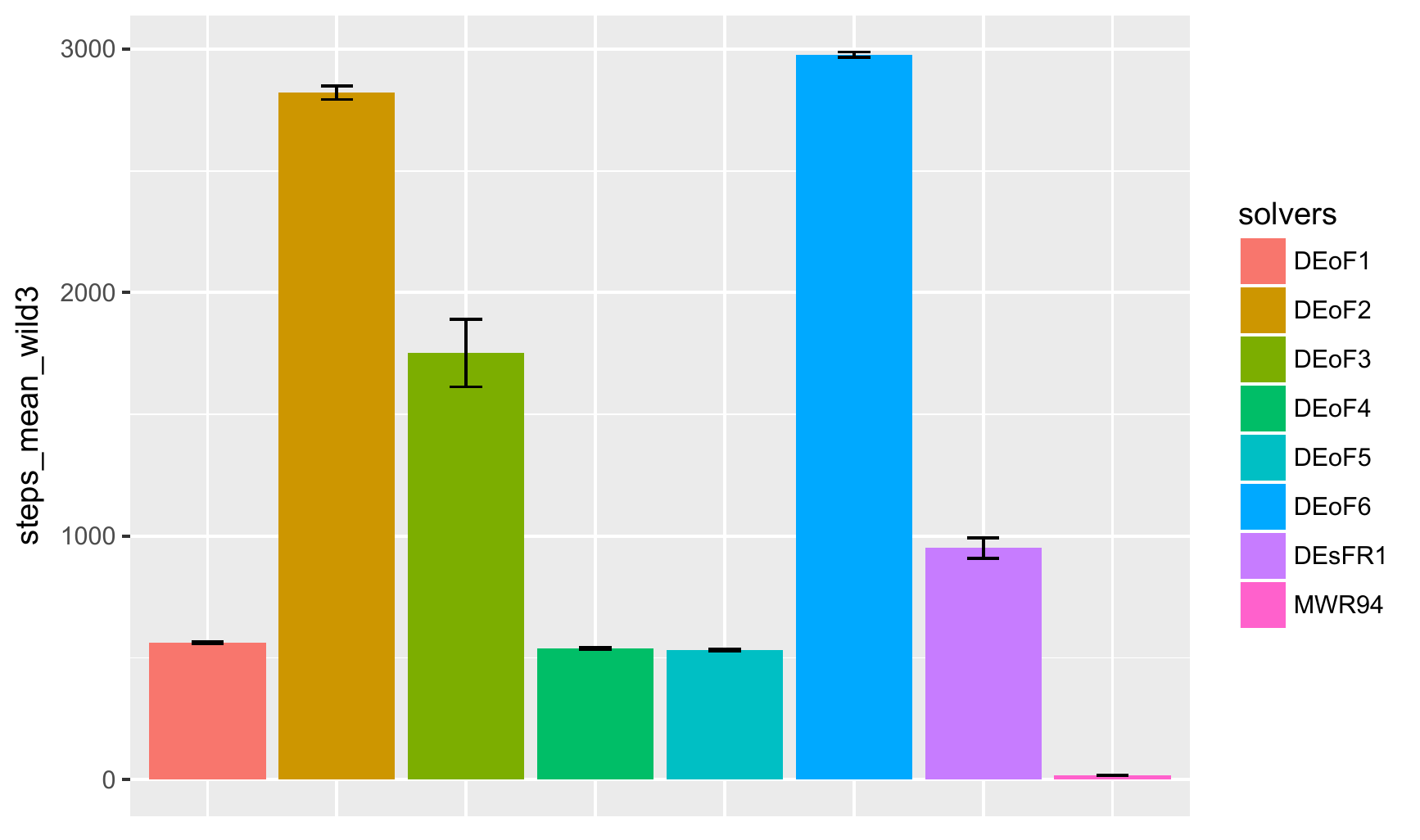}

\vspace*{2ex}
\caption{
Summary of first-passage-time experiments  with eight solvers and
two groups of three hard-to-solve functions: {\tt trefethen1}, {\tt trefethen2}, {\tt trefethen3}
and {\tt wild1}, {\tt wild2}, {\tt wild3}. 
The number of rulers associated with each function increases from 1 to 3.
The first seven solvers implement documented variants of 
the well-known DE algorithm described in the table under Figure~\ref{fg_mw_barplot_ehrenfest}.
The last solver, {\tt MWR}, is described in Section~\ref{sec_MWA} and Figure~\ref{fg_mw_stopping_templates}.
}
\label{fg_mw_barplot_hard}
\vspace*{-0ex}
\end{figure*}

\noindent
For a summary of first-passage-time experiments  with eight solvers and
two groups of three hard-to-solve functions, see Figure~\ref{fg_mw_barplot_hard}.
The number of rulers associated with each function increases from 1 to 3.
Function {\tt wild1} is from~\cite{OPUS-R-2011-JSS-Mullen}, 
functions {\tt trefethen2}, {\tt trefethen3} are from~\cite{OPUS-opt-2004-SIAM-Bornemann}.

\newpage

\section{Summary and Future Work}
\label{sec_summary}
\noindent
We expect to observe consistent and improved rate of convergence 
with  MW-solvers
also for other hard test instances
in continuous domain.
As we increase the neighborhood radius,
the increasing cost of computing the neighborhood matrix  
can be balanced with a parallel implementation.

An adaptation of multiwalk concepts to hard problems in discrete domains
will likely accelerate the convergence rate 
in comparison with  the current
state-of-the-art stochastic solvers such as reported in
\cite{OPUS2-2013-walk-MIDEM-Brglez}, 
\cite{OPUS2-2017-labs-Elsevier-Boskovic}, 
and \cite{OPUS2-2017-ogr-CEC-Brglez}.

\par\vspace*{1.9ex}\noindent
{\bf Acknowledgements.}
The solver DEoptim~\cite{OPUS-R-2011-JSS-Mullen} provided a robust foundation which we could not do without.
The elegance of simpleDE under~\cite{R-package-2018-adagio-Borchers} is  a model for our implementation. 
Suggestions  from Dr. Larry Nevin are much appreciated.


\end{multicols}

\end{document}